\documentclass{article}

\usepackage{microtype}
\usepackage{graphicx}
\usepackage{subfigure}
\usepackage{booktabs} 
\usepackage{multirow}
\usepackage{amsmath}
\usepackage{amsfonts}

\usepackage{hyperref}

\hypersetup{
    colorlinks=true,
    linkcolor=blue,
    filecolor=magenta,      
    urlcolor=cyan,
}

\newcommand{\iid}{\stackrel{\mathrm{i.i.d.}}{\sim}}

\usepackage[accepted]{icml2020}

\icmltitlerunning{Video Prediction via Example Guidance}

\begin{document}

\twocolumn[
\icmltitle{Video Prediction via Example Guidance}



\icmlsetsymbol{equal}{*}
\icmlsetsymbol{visit}{$\ddagger$}

\begin{icmlauthorlist}
\icmlauthor{Jingwei Xu}{equal,visit,sjtu}
\icmlauthor{Huazhe Xu}{equal,ucb}
\icmlauthor{Bingbing Ni}{sjtu}
\icmlauthor{Xiaokang Yang}{sjtu}
\icmlauthor{Trevor Darrell}{ucb}
\end{icmlauthorlist}

\icmlaffiliation{sjtu}{Shanghai Jiao Tong University}
\icmlaffiliation{ucb}{University of California, Berkeley}

\icmlcorrespondingauthor{Bingbing Ni}{nibingbing@sjtu.edu.cn}

\icmlkeywords{Multi-modality, Video Prediction}

\vskip 0.3in
]



\printAffiliationsAndNotice{\icmlEqualContribution. $^{\ddagger}$Work done during visiting Berkeley AI Research.}

\begin{abstract}
In video prediction tasks, one major challenge is to capture the multi-modal nature of future contents and dynamics.
In this work, we propose a simple yet effective framework that can efficiently predict plausible future states.
The key insight is that the potential distribution of a sequence could be approximated with analogous ones in a repertoire of training pool, namely, expert examples.
By further incorporating a novel optimization scheme into the training procedure, plausible predictions can be sampled efficiently from distribution constructed from the retrieved examples. 
Meanwhile, our method could be seamlessly integrated with existing stochastic predictive models; significant enhancement is observed with comprehensive experiments in both quantitative and qualitative aspects.
We also demonstrate the generalization ability to predict the motion of unseen class, i.e., without access to corresponding data during training phase.
Project Page: \hyperlink{https://sites.google.com/view/vpeg-supp/home.}{https://sites.google.com/view/vpeg-supp/home.}
\end{abstract}

\section{Introduction}
\label{sec:intro}
Video prediction involves accurately generating possible forthcoming frames in a pixel-wise manner given several preceding images as inputs.
As a natural routine for understanding the dynamic pattern of real-world motion, it facilitates many promising downstream applications, e.g., robot control, automatous driving and model-based reinforcement learning~\citep{DBLP:conf/nips/KurutachTYRA18,DBLP:conf/nips/NairPDBLL18,DBLP:conf/icml/PathakAED17}.

\begin{figure}[t]
\begin{center}
\centerline{\includegraphics[width=\columnwidth]{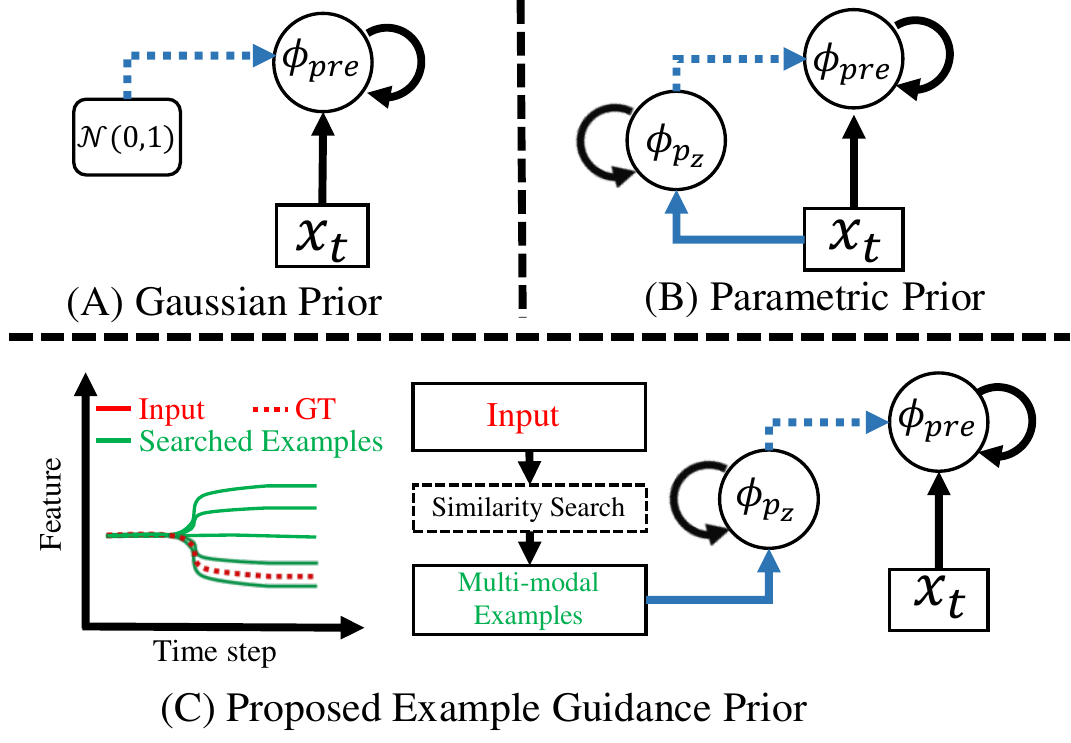}}
\vspace{-0.3cm}
\caption{Illustration of stochastic prediction with different prior schemes. Rectangle box refers to input. $\phi_{q_{z}}$ is for uncertainty modelling and $\phi_{pre}$ is the prediction model. We omit the output part for simplicity. Blue line corresponds to stochastic modelling and dashed line is the sampling procedure of random variable. (A) Prediction with fixed Gaussian prior, which does not consider the temporal dependency between different time steps. (B) Prediction with parametric prior, which lacks explicit supervision signal for multi-modal future modelling. (C) Proposed prediction scheme with similar examples retrieved in training dataset. These examples are utilized construct an explicit multi-modal distribution target for the training of prediction model.}
\label{Fig:Moti}
\end{center}
\vspace{-1.0cm}
\end{figure}

\citet{DBLP:conf/icml/SrivastavaMS15} first proposes to predict simple digit motion with deep neural models.
Video frames are synthesized in a deterministic manner~\citep{DBLP:conf/nips/DentonB17}, which also suffers to achieve long-range and high-quality prediction, even with large model capacity~\citep{DBLP:conf/nips/FinnGL16}.
\citet{DBLP:conf/iclr/BabaeizadehFECL18} shows that the distribution of frames is a more important aspect that should be modelled.
Variational based methods (e.g., SVG~\cite{DBLP:conf/icml/DentonF18} and SAVP~\cite{DBLP:journals/corr/abs-1804-01523})
are naturally developed to achieve good performance on simple dynamics such as digit moving~\citep{DBLP:conf/icml/SrivastavaMS15} and robot arm manipulation~\cite{DBLP:conf/nips/FinnGL16}.

However, real-world motion commonly follows multi-modal distributions.
With the increase of motion diversity and complexity, variational inference with prior Gaussian distribution is insufficient to cover the wide spectrum of future possibilities.
Meanwhile, downstream tasks mentioned in the first paragraph require prediction model with capability to model real-world distribution (i.e., can the multi-modal motion pattern be effectively captured?) and high sampling efficiency (i.e., fewer samples needed to achieve higher prediction accuracy).
These are both important factors for stochastic prediction, which are also the focus issues in this paper.
Recent work introduces external information (e.g., object location~\citep{ye2019cvp,DBLP:conf/icml/VillegasYZSLL17}) to ease the prediction procedure, which is hard to generalize to other scenes. 

Predictive models can heavily rely on similarity between past experiences and the new ones, implying that sequences with similar motion might fall into the same modal with a high probability.
The key insight of our work, deduced from the above observation, is that the potential distribution of sequence to be predicted can be approximated by analogous ones in a data pool, namely, examples.

In other words, our work (termed as \textbf{VPEG}, \textbf{V}ideo \textbf{P}rediction via \textbf{E}xample \textbf{G}uidance) bypasses implicit optimization of latent variable relying on variational inference; as shown in Fig.~\ref{Fig:Moti}C, we introduce an explicit distribution target constructed from analogous examples, which are empirically proved to be critical for distribution modelling.
To guarantee output predictions are multi-modal distributed, we further propose a novel optimization scheme which considers the prediction task as a stochastic process for explicit motion distribution modelling.
Meanwhile, we incorporate the adversarial training into proposed method to guarantee the plausibility of each predicted sample.
It is also worth mentioning that our model is able to integrate with the majority of existing stochastic predictive models.
Implementing our method is simply replacing variational method with the proposed optimization framework.
We conduct extensive experiments on several widely used datasets, including moving digit~\citep{DBLP:conf/icml/SrivastavaMS15}, robot arm motion~\citep{DBLP:conf/nips/FinnGL16}, and human activity~\citep{DBLP:conf/iccv/ZhangZD13}.
Considerable enhancement is observed both in quantitative and qualitative aspects.
Qualitatively, the high-level semantic structure, e.g., human skeleton topology, could be well preserved during prediction.  
Quantitatively, our model is able to produce realistic and accurate motion with fewer samples compared to previous methods.
Moreover, our model demonstrates generalization ability to predict unseen motion class during testing procedure, which suggests the effectiveness of example guidance.

\section{Related Work}
\textbf{Distribution Modelling with Stochastic Process.}
In this filed, one major direction is based on Gaussian process (denoted as GP) ~\citep{DBLP:books/lib/RasmussenW06}.
\citet{DBLP:conf/nips/WangFH05} proposes to extend basic GP model with dynamic formation, which demonstrates appealing ability of learning human motion diversity.
Another promising branch is determinantal point process (denoted as DPP)~\citep{DBLP:conf/icml/AffandiFAT14,DBLP:conf/icml/ElfekiCRE19}, which focuses on diversity of modelled distribution by incorporating a penalty term during optimization procedure.
Recently, the combination of stochastic process and deep neural network, e.g., neural process~\citep{DBLP:conf/icml/GarneloRMRSSTRE18} leads to a new routine towards applying stochastic process on large-scale data.
Neural process~\citep{DBLP:conf/icml/GarneloRMRSSTRE18} combines the best of both worlds between stochastic process (data-driven uncertainty modelling) and deep model (end-to-end training with large-scale data).
Our work, which treads on a similar path, focuses on the distribution modelling of real-world motion sequences.

\textbf{Video Prediction.} 
Video prediction is initially considered as a deterministic task which requires a single output at a time~\citep{DBLP:conf/icml/SrivastavaMS15}.
Hence, many works focus on the architecture optimization of the predictive models.
Conv-LSTM based model~\citep{DBLP:conf/nips/ShiCWYWW15,DBLP:conf/nips/FinnGL16,DBLP:conf/nips/WangLWGY17,DBLP:conf/cvpr/XuNLCY18,DBLP:conf/iclr/LotterKC17,DBLP:conf/eccv/ByeonWSK18,DBLP:conf/iclr/WangJYLLF19} is then proposed to enhance the spatial-temporal connection within latent feature space to pursue better visual quality.
High fidelity prediction could be achieved by larger model and more computation sources~\cite{DBLP:conf/nips/VillegasPKELL19}.
Flow-based prediction model~\cite{DBLP:conf/iclr/KumarBEFLDK20} is proposed to increase the interpretability of the predicted results.
Disentangled representation learning~\citep{DBLP:conf/nips/DentonB17,DBLP:journals/corr/abs-1812-00452} is proposed to reduce the difficulty of human motion modelling~\cite{DBLP:conf/mm/YanXNZY17} and prediction.
Another branch of work~\citep{DBLP:conf/nips/JiaBTG16} attempts to predict the motion with dynamic network, where the deep model is flexibly configured according to inputs, i.e., adaptive prediction.
Deterministic model is infeasible to handle multiple possibilities.
Stochastic video prediction is then proposed to address this problem.
SV2P~\citep{DBLP:conf/iclr/BabaeizadehFECL18} is firstly proposed as an stochastic prediction framework incorporated with latent variables and variational inference for distribution modelling.
Following a similar inspiration, SAVP~\citep{DBLP:journals/corr/abs-1804-01523} demonstrates that the combination of GAN~\citep{DBLP:conf/nips/GoodfellowPMXWOCB14} and VAE~\citep{DBLP:journals/corr/KingmaW13} facilitates better modelling of the future possibilities and significantly boosts the generation quality of predicted frames.
\citet{DBLP:conf/icml/DentonF18} proposes to model the unknown true distribution in a parametric and learnable manner, i.e., represented by a simple LSTM~\citep{DBLP:journals/neco/HochreiterS97} network. 
Recently, unsupervised keypoint learning~\citep{DBLP:conf/nips/KimNCK19} (i.e., human pose~\cite{Xu_2020_CVPR}) is utilized to ease the modelling difficulty of future frames.
Domain knowledge, which helps to reduce the motion ambiguity~\citep{ye2019cvp,DBLP:conf/nips/TangS19,DBLP:conf/iccv/LucNCVL17}, is proved to be effective in future prediction.
In contrast to these works above, we are motivated by one insight that prediction is based on similarity between the current situation and the past experiences.
More specifically, we argue that the multi-modal distribution could be effectively approximated with analogous ones (i.e., examples) in training data and real-world motion could be further accurately predicted with high sampling efficiency.

\begin{figure}[t]
\begin{center}
\centerline{\includegraphics[width=\columnwidth]{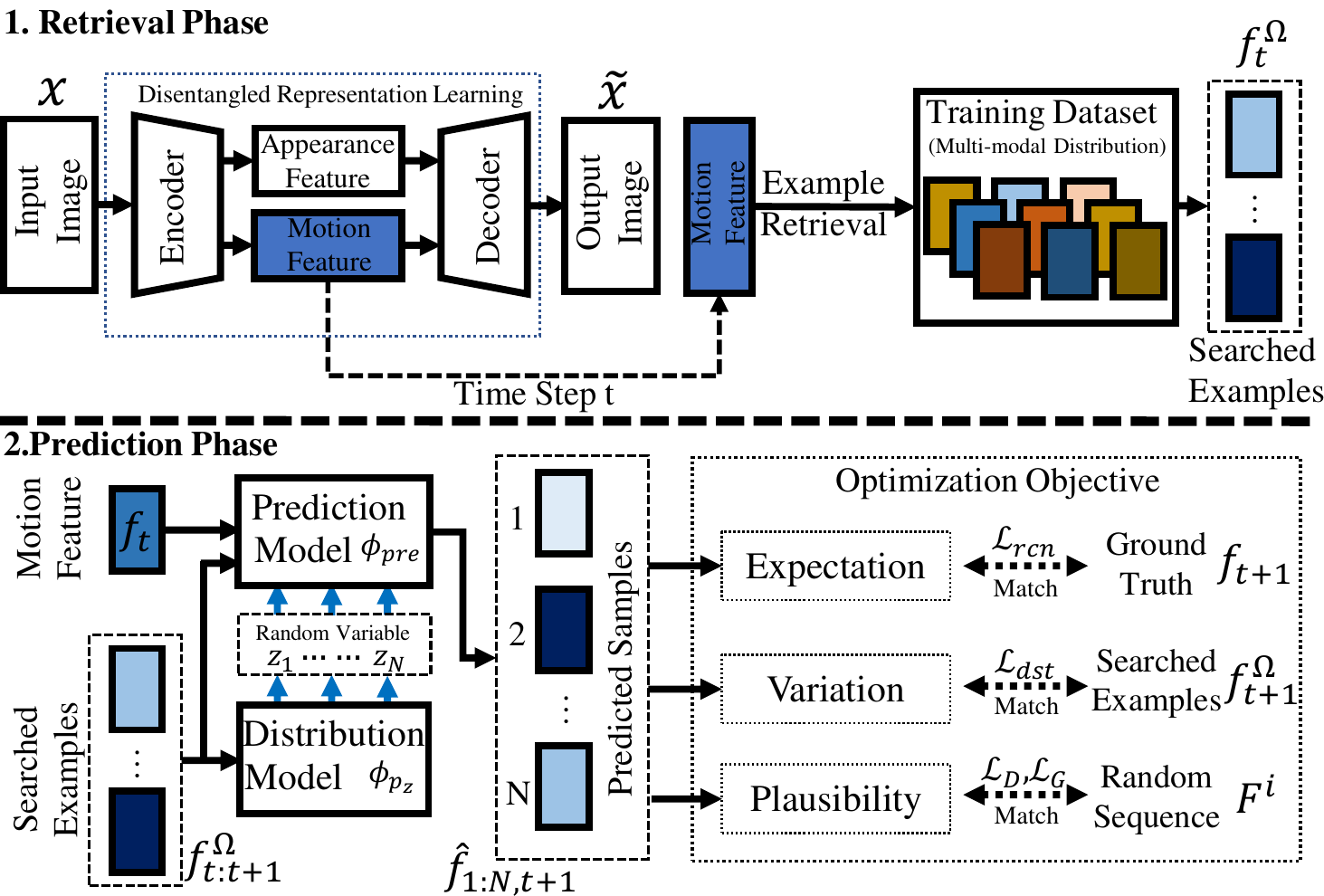}}
\vspace{-0.3cm}
\caption{Overall framework of proposed video prediction method. The whole procedure is split into two consecutive phases presented at the top and bottom rows respectively. Top row refers to retrieval process of proposed method, while bottom rows is the prediction model with example guidance. It is optimized as a stochastic process to effectively capture the future motion uncertainty.}
\label{Fig:framework}
\end{center}
\vspace{-1.0cm}
\end{figure}

\section{Method}
Given $M$ consecutive frames as inputs, we are to predict the future $N$ frames in the pixel-wise manner.
Suppose the input (context) frames $\mathbf{X}$ is of length $M$, i.e., $\mathbf{X}=\{\mathbf{x}_{t}\}_{t=1}^{M}\in \mathbb{R}^{W\times H \times C \times M}$, where $W,H,C$ are image width, height and channel respectively.
Following the notation defined, the prediction output $\mathbf{Y}$ is of length $N$, i.e., $\mathbf{Y}=\{\mathbf{y}_{t}\}_{t=1}^{N}\in \mathbb{R}^{W\times H \times C \times N}$.
We denote the whole training set as $\mathcal{D}_{s}$.
Fig.~\ref{Fig:framework} demonstrates the overall framework of the proposed method.
Details are presented in following subsections.

\begin{figure}[t]
\begin{center}
\centerline{\includegraphics[width=0.95\columnwidth]{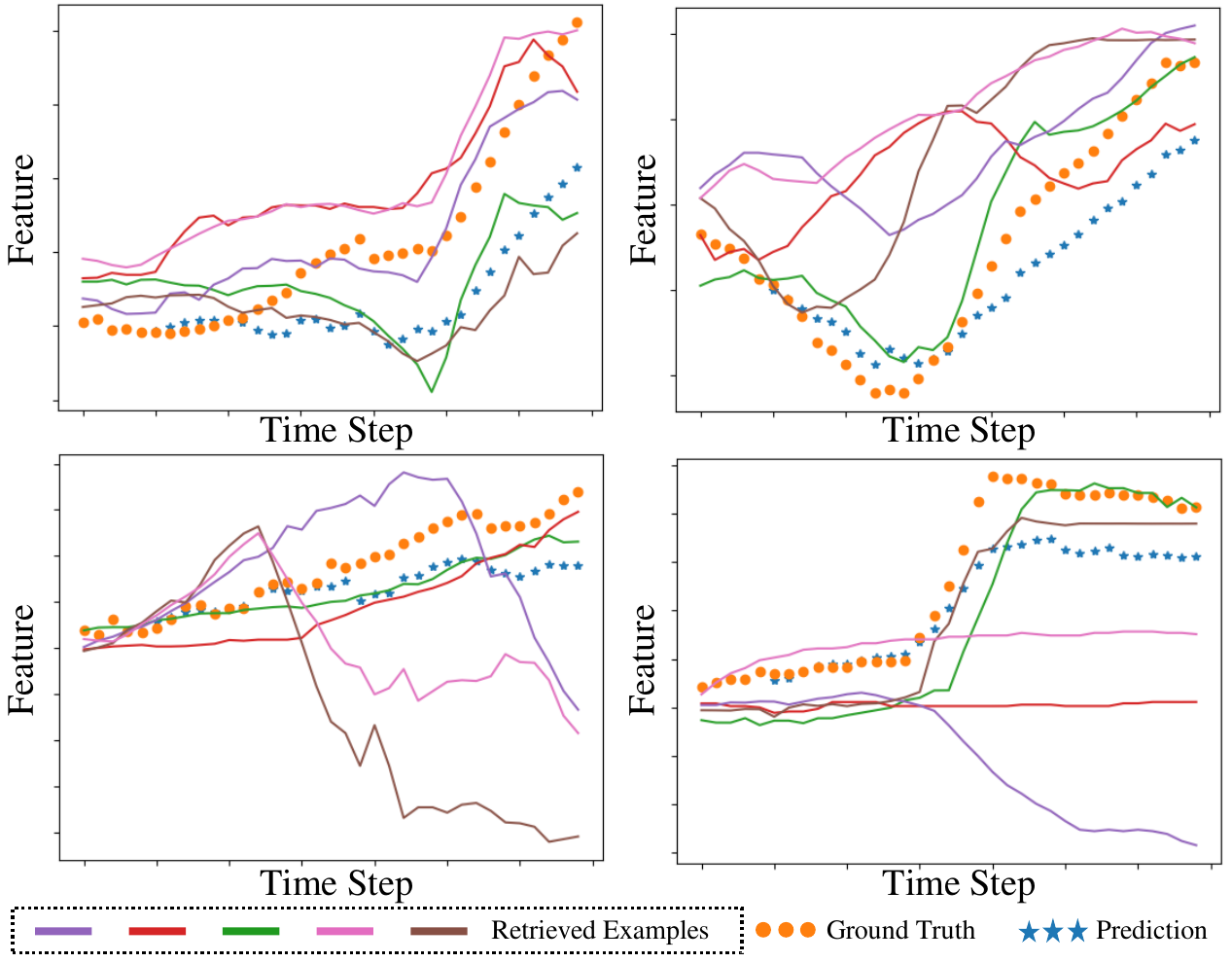}}
\vspace{-0.3cm}
\caption{Four typical patterns of retrieved examples on PennAction~\citep{DBLP:conf/iccv/ZhangZD13} dataset. The five solid lines refer to top-5 examples searched in $\mathcal{D}_{s}$ and orange-dot line is the ground truth motion sequence. The blue-star line is predicted sequence. The input sequence generally falls into one variation pattern of retrieved examples, which confirms the key insight of our work.}
\label{Fig:example}
\end{center}
\vspace{-1.0cm}
\end{figure}

\subsection{Example Retrieval via Disentangling Model}\label{Sec:Demo_Retriv}
We conduct the retrieval procedure in training set $\mathcal{D}_{s}$.
To avoid trivial solution, $\mathbf{X}$ is excluded from $\mathcal{D}_{s}$ if $\mathbf{X}$ is in the training set $\mathcal{D}_{s}$.
Direct search in the image space is infeasible because it generally contains unnecessary information for retrieval, e.g., the appearance of foreground subject and detailed structure of background.
Alternatively, a better solution is retrieving in disentangled latent space.
Many previous methods~\citep{DBLP:conf/nips/DentonB17,DBLP:conf/cvpr/Tulyakov0YK18,DBLP:conf/iclr/VillegasYHLL17,DBLP:conf/icml/DentonF18} have made promising progress in learning to disentangle latent feature.
Two competitive methods, i.e.,  SVG~\citep{DBLP:conf/icml/DentonF18} and~\citet{DBLP:conf/nips/KimNCK19} are adopted as the disentangling model in our work.
~\citet{DBLP:conf/nips/KimNCK19} proposes an unsupervised method to extract keypoints of arbitrary object, whose pretrained model is directly used to extract the pose information as motion feature in our work.
Note that the motion feature remains valid when input is only one frame, where the single state is treated as motion feature.
SVG~\citep{DBLP:conf/icml/DentonF18} unifies the disentangling model and variational inference based prediction into one stage.
We remove the prediction part and train the disentangling model as:
\begin{align}
(\mathbf{b}_{t},\mathbf{h}_{t})&=\phi_{dse}(\mathbf{x}_{t}),t\in \{i,j\},\\
  \mathcal{L}_{dse}&=||\phi_{dec}(\mathbf{b}_{i}, \mathbf{h}_{j})-\mathbf{x}_{j}||_{2}^{2},
\end{align}
where $i,j$ are two random time steps sampled from one sequence, $\phi_{dse}$ and $\phi_{dec}$ are disentangling model and decoder (for image reconstruction) respectively.
$\mathcal{L}_{dse}$ indicates two frames from the same sequence share similar background and should be able to reconstruct each other by exchanging the motion feature.
By optimizing this loss function, appearance feature $\mathbf{b}_{*}$ is expected to be constant while $\mathbf{h}_{*}$ contains the motion information, which leads the disentanglement model to learn to extract motion feature in a self-supervised way.
Both disentanglement models SVG~\citep{DBLP:conf/icml/DentonF18} and~\citet{DBLP:conf/nips/KimNCK19} could be presented in a unified way as shown in Fig.~\ref{Fig:framework}.
Next we focus on the retrieval procedure.
Note that all input frames are used in this part.
We denote the feature used for retrieval as $\mathbf{F}\in \mathbb{R}^{C_{f}\times M}=\{\mathbf{f}_{t}\}_{t=1}^{M}$, where $C_{f}$ is the number of feature dimension.

Given input sequence $\mathbf{X}$, whose motion feature denoted as $\mathbf{F}$, and training set $\mathcal{D}_{s}$, we conduct nearest-neighbor search as:
\begin{equation}
\label{Eqn:NN}
\Omega_{i}=\mathcal{S}(||\mathbf{F}^{i}-\mathbf{F}||_{2}^{2}, K),
\end{equation}
where $\mathbf{F}^{i}$ refers to the extracted feature of $\mathbf{X}^{i}\in \mathcal{D}_{s}$. $\mathcal{S}(\bullet, K)$ refers to top $K$ selection from a set in the ascending order.
$\Omega_{i}$ is the retrieved index set corresponding to $i$th sample.
Note that the subscript $i$ is omitted for simplicity in following contexts.
$K$ is treated as a hyper-parameter in our experiments, whose influence is validated through ablation study in Sec.~\ref{Sec:Exp}.
We perform first-order difference along the temporal axis to focus on state difference during the retrieval procedure if multiple frames are available.
We plot retrieved examples (solid line, $K=5$) together with the input sequence (orange-dot line) in Fig.~\ref{Fig:example}.
Here we have two main observations:
(1) The input sequence generally falls into one motion pattern of retrieved examples, which confirms the key insight of our work.
(2) The examples have non-Gaussian distribution, which implies the difficulty on the optimization side by a variational inference method.
We present more visual evidence in supplementary material to demonstrate the common existence of such similarity between $\mathbf{F}$ and $\mathbf{F}^{\Omega}$.

\textbf{Discussion of Retrieval Efficiency.}
Note that the retrieval module is introduced in this work, which is an additional step compared to the majority of previous methods.
One concern would thus be the retrieval time, which is highly correlated with efficiency of the whole model.
We would like to clarify that the retrieval step is highly efficient: (1) It is executed in the low dimensional feature space (i.e., $\mathbf{f}\in\mathbb{R}^{C_{f}}$) rather than in the image space, which requires less computation; (2) It is implemented with the efficient quick-sort algorithm. The averaged retrieval complexity is O(NlogN), where N is the number of video sequences.
For example, on the PennAction dataset~\cite{DBLP:conf/iccv/ZhangZD13} (containing 1172 sequences in total) the whole running (including retrieval) time of predicting 32 frames is 354ms, while the retrieval time only takes 80ms.

\subsection{Example Guided Multi-modal Prediction}
\label{Sec:VPDG}
\subsubsection{Stochastic Video Prediction Revisited}
The majority works~\citep{DBLP:conf/nips/DentonB17,ye2019cvp,DBLP:conf/icml/DentonF18,DBLP:journals/corr/abs-1804-01523} of stochastic video prediction are based on variational inference. 
We first briefly review previous works in this field and then analyze the inferiority of stochastic prediction based on variational inference.

These methods use a latent variable (denoted as $\mathbf{z}$) to model the future uncertainty.
The distribution of $\mathbf{z}$ (denoted as $p_{z}$) is trained to match with a (possibly fixed) prior distribution (denoted as $q_{z}$) as follows,
\begin{equation}
\mathcal{L}=||\phi_{pre}(\mathbf{F}_{1:t-1},\mathbf{z}_{t})-\mathbf{f}_{t}||_{2}^{2}+\mathcal{L}_{KL}(p_{z}||q_{z}),
\end{equation}
where $\phi_{pre}$, $q_{z}$, $\mathcal{L}_{KL}()$ are the prediction model, target distribution and Kullback-Leibler divergence function~\citep{Kullback51klDivergence} respectively.
$p_{z}$ is generally modelled with deep neural network (e.g., $\phi_{p_{z}}(\mathbf{f}_{t-1})$).
$q_{z}$ is fixed, e.g., $\mathcal{N}(\mathbf{0},\mathbf{I})$.
This implies that the predicted image $\mathbf{X}_{t}$ is controlled by $\mathcal{N}(\mathbf{0},\mathbf{I})$, not real-world motion distribution.
\citet{DBLP:conf/icml/DentonF18} proposes to model the potential distribution with $\phi_{q_{z}}(\mathbf{f}^{t})$, which still lacks an explicit supervision signal on the distribution of motion feature.

The essence of the modelling difficulty is from the optimization target of the $\mathcal{L}_{KL}$ term.
Under the framework of variational inference, the form of $q_{z}$ is generally restricted to a normal distribution for tractability.
However, this is in conflict with the multi-modal distribution nature of real-world motion.
We need a more explicit and reliable target and thus propose to construct it with similar examples $\mathbf{f}^{\Omega}$ whose retrieval procedure is described in Sec.~\ref{Sec:Demo_Retriv}.

\begin{algorithm}[tb]
   \caption{Example Guided Video Prediction}
   \label{Alg:DGVP}
\begin{algorithmic}
   \STATE {\bfseries Input:} Training Set $\mathcal{D}_{s}$, disentangling model $\phi_{dse}$, predictor $\phi_{pre}$ and discriminator $\phi_{dcm}$.
   
   $\#$\emph{Example retrieval phase}
   \FOR{Input sequence $\mathbf{X}$ in $\mathcal{D}_{s}$}
   \STATE Get motion feature $\mathbf{F}=\phi_{dse}(\mathbf{X})$,
   \STATE Example retrieval to obtain $\mathbf{F}^{\Omega}$ as Eqn.~\ref{Eqn:NN}.
   \ENDFOR
   
   $\#$\emph{Prediction phase}
   \REPEAT
   \STATE Get a random batch of $(\mathbf{F},\mathbf{F}^{\Omega})$ pairs.
   
   $\#$\emph{Optimization as a stochastic process}
   \FOR{$i=1$ {\bfseries to} $N$}
   \STATE Sample noise $\mathbf{z}_{i,t+1}$ as Eqn.~\ref{Eqn:noise},
   \STATE Predict next state $\hat{\mathbf{f}}_{i,t+1}$ as Eqn.~\ref{Eqn:predict},
   \ENDFOR
   \STATE Optimize w.r.t. Eqn.~\ref{Eqn:rcn},~\ref{Eqn:dst},~\ref{Eqn:ld} and~\ref{Eqn:lg}.
   \UNTIL{the training objective $\mathcal{L}_{fin}$ (Eqn.~\ref{Eqn:final}) converged.}
\end{algorithmic}
\end{algorithm}

\subsubsection{Prediction with Examples}
Given retrieved examples $\mathbf{f}^{\Omega}$, we first construct a new distribution target and then learn to approximate it.
The most straightforward way is directly replacing the prior distribution $q_{z}$ with the new one.
More specifically, at time step $t$ the distribution model $\phi_{p_{z}}$ is trained as:
\begin{equation}
\hat{\mu}_{t},\hat{\sigma}_{t}=\phi_{p_{z}}(\hat{\mathbf{F}}_{1:t-1}),\mathbf{z}_{t}\sim\mathcal{N}(\hat{\mu}_{t},\hat{\sigma}_{t}),
\end{equation}
\begin{equation}
\mu_{t},\sigma_{t} = \phi_{q_{z}}(\mathbf{f}^{\Omega}_{t}),\mathcal{L}_{KL}=\log(\frac{\sigma_{t}}{\hat{\sigma}_{t}})+\frac{\hat{\sigma}_{t}+(\mu_{t}-\hat{\mu}_{t})^{2}}{2\sigma_{t}},
\end{equation}
where $\mathbf{z}_{t}$ models the possibility of future state and $\mu_{t},\sigma_{t}$ are commonly supervised with $\mathcal{L}_{KL}$.

However, it is difficult to obtain promising results with the above method which simply replaces $\mathbf{f}_{t}$ with $\mathbf{f}^{\Omega}_{t}$.
The reason mainly lies in two aspects: 
Firstly, the diversity of predicted motion feature at time step $t$ (denoted as $\hat{\mathbf{f}}_{t}$) lacks an explicit supervision signal.
Secondly, the distribution of latent variable $\mathbf{z}_{t}$ (i.e., $\mathcal{N}(\mu_{t},\sigma_{t})$) is infeasible to accurately represent the motion diversity of $\mathbf{f}_{t}^{\Omega}$, because no dedicated training objective is designed for this target.

\textbf{Optimization as Stochastic Process.}
Motivated by the above two issues, we consider the prediction task as a stochastic process targeting at explicit distribution modelling.
The whole prediction procedure is conducted in motion feature space.
The inputs of prediction model $\phi_{pre}$ include $\mathbf{f}^{\Omega}_{t}$ and $\mathbf{f}_{t}$.
We calculate the mean and variance of example feature $\mathbf{f}^{\Omega}_{t}$, i.e., $\mathcal{E}(\mathbf{f}^{\Omega}_{t})$ and $\mathcal{V}(\mathbf{f}^{}_{t})$, for the subsequent random sampling in motion space.
The prediction procedure at time step $t$ is conducted as follows,
\begin{equation}
\label{Eqn:randz}
(\mu_{t},\sigma_{t})=\phi_{q_{z}}(\mathcal{E}(\mathbf{f}_{t}^{\Omega}),\mathcal{V}(\mathbf{f}_{t}^{\Omega})),
\end{equation}
\begin{equation}
\label{Eqn:noise}
(\mathbf{z}_{1,t},...,\mathbf{z}_{N,t})\iid\mathcal{N}(\mu_{t},\sigma_{t}),
\end{equation}
\begin{equation}
\label{Eqn:predict}
\hat{\mathbf{f}}_{i,t+1}=\phi_{pre}(\hat{\mathbf{f}}_{i,t},\mathbf{z}_{i,t},\mathbf{f}_{t}^{\Omega}),i=1,...,N,
\end{equation}
where $(\mathbf{z}_{1,t},...,\mathbf{z}_{N,t})$ is a group of independent and identically sampled values and $\mathbf{z}_{i,t}\in \mathbb{R}^{h}$.
The subscript ${i,t}$ refers to $i$th sample at time step $t$.
Predicted state $\hat{\mathbf{f}}_{i,t}$ is not fed into $\phi_{q_{z}}$, where we empirically get sub-optimal results.
Because at initial training stage $\hat{\mathbf{f}}_{i,t}$ is noisy and non-informative, which in turn acts as a distractor for training $\phi_{q_{z}}$.
The prediction model is trained as follows,

\begin{equation}
\label{Eqn:rcn}
\mathcal{L}_{rcn}=||\hat{\mathbf{f}}_{j,t+1}-\mathbf{f}_{t+1}||_{2}^{2},
\end{equation}
\begin{equation}
\label{Eqn:dst}
\mathcal{L}_{dst}=||\mathcal{V}(\{\hat{\mathbf{f}}_{i,t+1}\}_{i=1}^{N})-\mathcal{V}(\mathbf{f}_{t+1}^{\Omega})||_{2}^{2},
\end{equation}
where $j=\min_{i}||\{\hat{\mathbf{f}}_{i,t+1}\}_{i=1}^{N}-\mathbf{f}_{t+1}||_{2}^{2}$. 
$\mathcal{L}_{rcn}$ indicates that the best matched one is used for training~\cite{DBLP:conf/nips/XuNY18}. Empirically, it is proved to be useful for stabilizing prediction when having multiple outputs.

$\mathcal{L}_{dst}$ aims to restrict the variety of $N$ predicted features to match with $\mathbf{f}_{t+1}^{\Omega}$.
In this way, the motion information of examples is effectively utilized and the distribution of predicted sequences is explicitly supervised.
Meanwhile, to guarantee the plausibility of each predicted sequence, we incorporate the adversarial training into our method. 
More specifically, a motion discriminator $\phi_{dcm}$ is utilized to facilitate realistic prediction.
\begin{equation}
\label{Eqn:ld}
\mathcal{L}_{D}=\frac{1}{2}(\phi_{dcm}(1-\mathbf{F})+\phi_{dcm}(1+\hat{\mathbf{F}}_{i})),
\end{equation}
\begin{equation}
\label{Eqn:lg}
\mathcal{L}_{G}=-\phi_{dcm}(1-\hat{\mathbf{F}}_{i}),
\end{equation}
where $i\in [1,N]$, $\mathcal{L}_{D},\mathcal{L}_{G}$ are adversarial losses for $\phi_{dcm}$ and $\phi_{pre}$~\citep{DBLP:conf/nips/GoodfellowPMXWOCB14}.
Adversarial training effectively guarantees the predicted sequence not drifting far away from the real-wold motion examples.
For clarity we present the whole prediction procedure in Alg.~\ref{Alg:DGVP}.

\textbf{Improvement upon existing models.}
Our work mainly focuses on multi-modal distribution modelling and sampling efficiency, which is adaptive to multiple neural models.
In Sec.~\ref{Sec:Exp}, we demonstrate extensive results by combining proposed framework with two baselines, i.e., SVG~\cite{DBLP:conf/icml/DentonF18} and~\citet{DBLP:conf/nips/KimNCK19}. The final objective is shown below:
\begin{equation}
\mathcal{L}_{fin}=\lambda_{1}\mathcal{L}_{rcn}+\lambda_{2}\mathcal{L}_{dst}+\lambda_{3}\mathcal{L}_{D}+\lambda_{4}\mathcal{L}_{G}.
\label{Eqn:final}
\end{equation}
For training and implementation details, (hyper-parameter and network architecture), please refer to the supplementary material.

\begin{figure}[t]
\begin{center}
\centerline{\includegraphics[width=1.0\columnwidth]{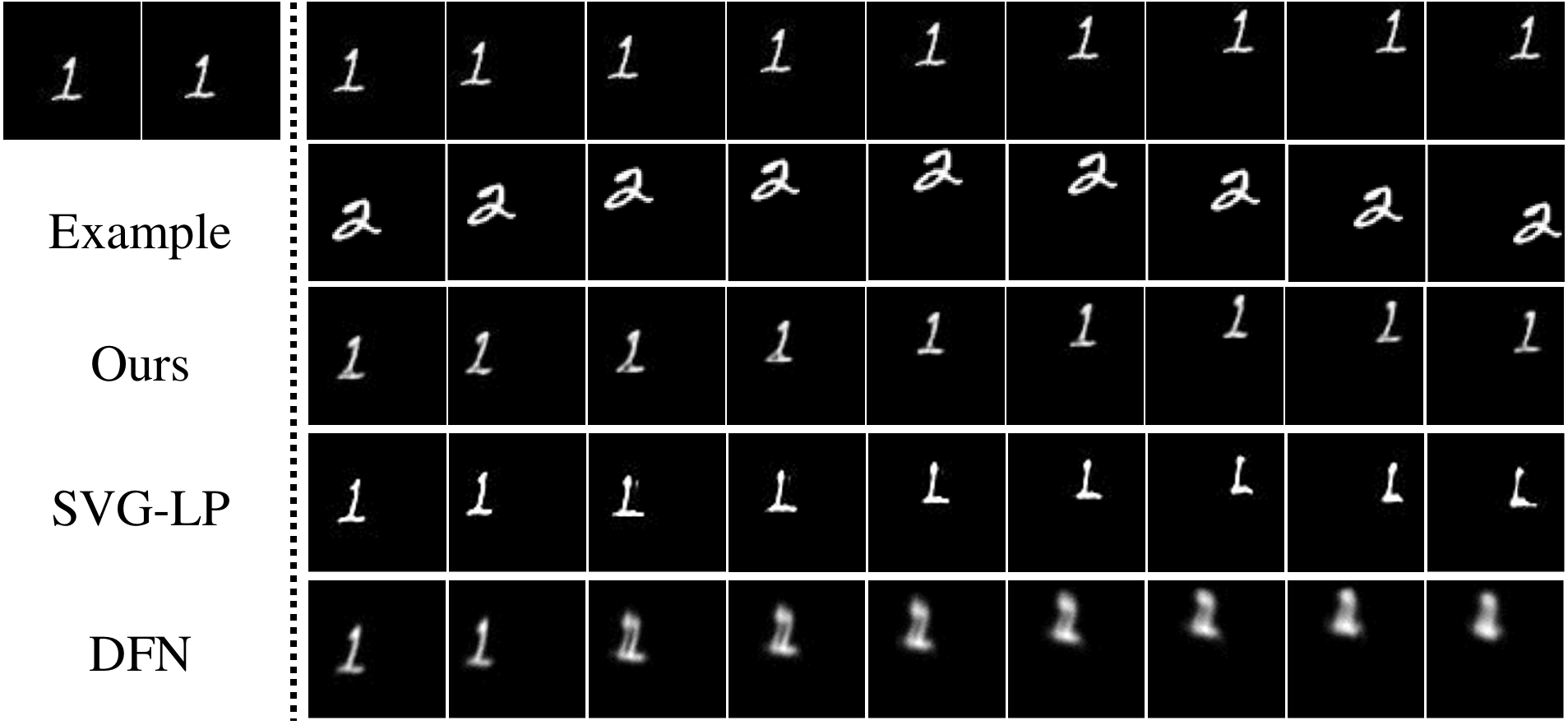}}
\vspace{-0.3cm}
\caption{Visualization of prediction results on MovingMnist~\citep{DBLP:conf/icml/SrivastavaMS15} dataset under stochastic setting. First row refers to ground truth. Following three rows correspond to example, predicted sequences of proposed model, SVG~\citep{DBLP:conf/icml/WichersVEL18} and DFN~\citep{DBLP:conf/nips/ShiCWYWW15} respectively.}
\label{Fig:MotiExp}
\end{center}
\vspace{-1.0cm}
\end{figure}

\section{Experiments}
\label{Sec:Exp}
\subsection{Datasets and Evaluation Metrics}
We evaluate our model with three widely used video prediction datasets: (1) MovingMnist~\citep{DBLP:conf/icml/SrivastavaMS15}, (2) Bair RobotPush~\citep{DBLP:conf/corl/EbertFLL17} and (3) PennAction~\citep{DBLP:conf/iccv/ZhangZD13}.
Following the evaluation practice of SVG~\citep{DBLP:conf/iclr/BabaeizadehFECL18} and~\citet{DBLP:conf/nips/KimNCK19}, we calculate the per-step prediction accuracy in terms of PSNR and SSIM.
The overall prediction quality of video frames is evaluated with Fréchet Video Distance (FVD)~\citep{DBLP:journals/corr/abs-1812-01717}.
To ensure fair evaluation, we compare with models whose source code is publicly available.
Specifically, on MovingMnist~\citep{DBLP:conf/icml/SrivastavaMS15} dataset we compare with SVG~\citep{DBLP:conf/icml/DentonF18} and DFN~\citep{DBLP:conf/nips/ShiCWYWW15}; On RobotPush~\citep{DBLP:conf/corl/EbertFLL17} dataset SVG~\citep{DBLP:conf/icml/DentonF18}, SV2P~\citep{DBLP:conf/iclr/BabaeizadehFECL18} and CDNA~\citep{DBLP:conf/nips/FinnGL16} are treated as baselines; On PennAction~\citep{DBLP:conf/iccv/ZhangZD13} dataset the works of~\citet{DBLP:conf/nips/KimNCK19,DBLP:conf/eccv/LiFYWLY18,DBLP:conf/icml/WichersVEL18,DBLP:conf/icml/VillegasYZSLL17} are used for comparison.
Note that to follow the best practice of the baseline model~\cite{DBLP:conf/nips/KimNCK19}, the prediction procedure on the PennAction Dataset~\cite{DBLP:conf/iccv/ZhangZD13} is a implementation-wise variant of  Eqn.~\ref{Eqn:randz}-\ref{Eqn:predict}. 
More specifically, the random noise is sampled only at the first time stamp.
Please refer to prediction procedure of the baseline model~\cite{DBLP:conf/nips/KimNCK19} for more details.
In all experiments we empirically set $N=K=5$.

\begin{table}[t]
	\centering
	\small
	\tabcolsep=1mm
	\resizebox{\linewidth}{!}{
		\begin{tabular}{c|l|@{}rrrrrrrrrr@{}}
		\toprule
		\toprule
		\textbf{Mode} & \textbf{Model} & T=1 & T=3 &  T=5 & T=7 & T=9 & T=11 & T=13 & T=15 & T=17\\
		\midrule
		\multirow{3}{*}{D} & DFN & 25.3 & 23.8 & 22.9 & 22.0 & 21.2 & 20.1 & 19.5 & 19.1 & 18.9 \\
		& SVG-LP & 24.7 & 22.8 & 21.3 & 19.5 & 18.8 & 18.2  & 17.9 & 17.7 & 17.4 \\
		& Ours & 25.6 & 23.2 & 22.5 & 21.7 & 20.8 & 20.3 & 19.8 & 19.5 & 19.3 \\
		\midrule
		\multirow{3}{*}{S} & DFN & 25.1 & 22.1 & 18.9 & 16.5 & 16.2 & 15.7 & 15.2  & 14.9 & 14.3 \\
		& SVG-LP & 25.4 & 23.9 & 22.9 & 19.5 & 19.0 & 18.7  & 18.7 & 18.2 & 17.6 \\
		& Ours & 26.0 & 24.8 & 23.1 & 22.1 & 21.0 & 20.5 & 19.7 & 19.5 & 19.2 \\
		\bottomrule
		\bottomrule
		\end{tabular}
	}
	\vspace{-0.3cm}
	\caption{Prediction accuracy on MovinMnist dataset~\cite{DBLP:conf/icml/SrivastavaMS15} in terms of PSNR. Mode refers to experiment setting, i.e., stochastic (S) or deterministic (D). We compare our model with SVG-LP~\citep{DBLP:conf/icml/DentonF18} and DFN~\citep{DBLP:conf/nips/JiaBTG16}.}
	\vspace{-0.5cm}
	\label{Tbl:PSNR_Mnist}
\end{table}

\subsection{Motivating Experiments: Moving Digit Prediction}
For MovingMnist~\citep{DBLP:conf/icml/SrivastavaMS15} dataset, inputs/outputs are of length 5 and 10 respectively during training.
Note that this dataset is configured with two different settings, i.e.,  to be deterministic or stochastic.
The deterministic version implies that the motion is determined by initial direction and velocity, while for the stochastic one, a new direction and velocity are applied after the digit hitting the boundary.
\emph{The prediction model should be able to accurately estimate motion patterns under both settings.} 

\textbf{Deterministic Motion Prediction.}
Tab.~\ref{Tbl:PSNR_Mnist} shows prediction accuracy (in terms of PSNR) from  T=1 to T=17.
One can observe that our model outperforms SVG-LP~\citep{DBLP:conf/icml/DentonF18} by a large margin and is comparable to DFN~\citep{DBLP:conf/nips/JiaBTG16}.
Under deterministic setting the retrieved examples provide exact motion information to facilitate prediction procedure.
We present corresponding visual results in supplementary material and please refer to it.

\begin{figure}[t]
\begin{center}
\centerline{\includegraphics[width=\columnwidth]{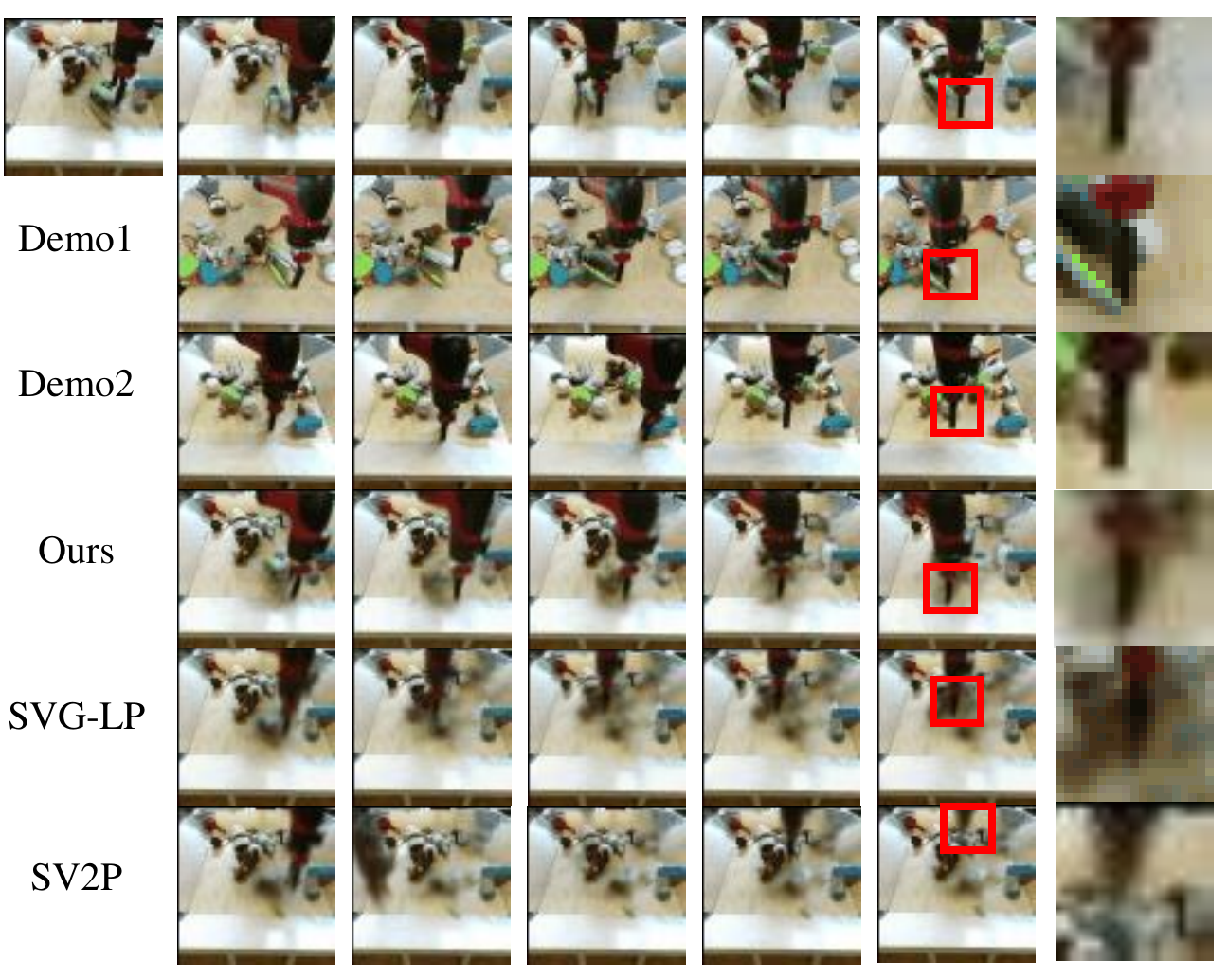}}
\vspace{-0.3cm}
\caption{Comparison of the predicted sequences on RobotPush~\citep{DBLP:conf/corl/EbertFLL17} dataset. Rows from top to bottom: ground truth, two retrieved examples, predicted results of our model, SVG~\citep{DBLP:conf/icml/DentonF18} and SV2P~\citep{DBLP:conf/iclr/BabaeizadehFECL18}.}
\label{Fig:RobotQual}
\end{center}
\vspace{-0.5cm}
\end{figure}

\textbf{Stochastic Motion Prediction.}
Under stochastic setting, the best PSNR value of 20 random samples is reported (bottom three rows of Tab.~\ref{Tbl:PSNR_Mnist}).
Considerable improvement over SVG-LP~\citep{DBLP:conf/icml/DentonF18} could be observed from Tab.~\ref{Tbl:PSNR_Mnist}.
Despite the retrieved example sequence not perfectly matching with ground truth (Fig.~\ref{Fig:MotiExp}, first two columns refer to input.), informative motion pattern is provided, i.e., bouncing back after reaching the boundary.
The deterministic model (DFN~\citep{DBLP:conf/nips/JiaBTG16}), which only produces a single output, is infeasible to properly handle stochastic motion.
For example, the blur effect (last row in Fig.~\ref{Fig:MotiExp}) is observed after hitting the boundary.

Deterministic and stochastic datasets possess different motion patterns and distributions. 
Non-stochastic method (e.g., DFN~\citep{DBLP:conf/nips/JiaBTG16}) is insufficient to capture motion uncertainty, while SVG-LP~\citep{DBLP:conf/icml/DentonF18}, empirically restricted by the stochastic prior nature in variational inference, is not capable of accurately predicting the trajectory under the deterministic condition.
Under deterministic setting, retrieved examples generally follow similar trajectory, whose variance is low.
For the stochastic version, searched sequences are highly diverse but follow the same motion pattern, i.e., bouncing back when hitting the boundary.
Guided by examples from these experiences, our model is able to reliably capture the motion pattern under both settings.
It implies that compared to fixed/learned prior, the motion variety could be better represented by similar examples.
Please refer to supplementary material for more visual results.

\begin{figure}[t]
\begin{center}
\centerline{\includegraphics[width=\columnwidth]{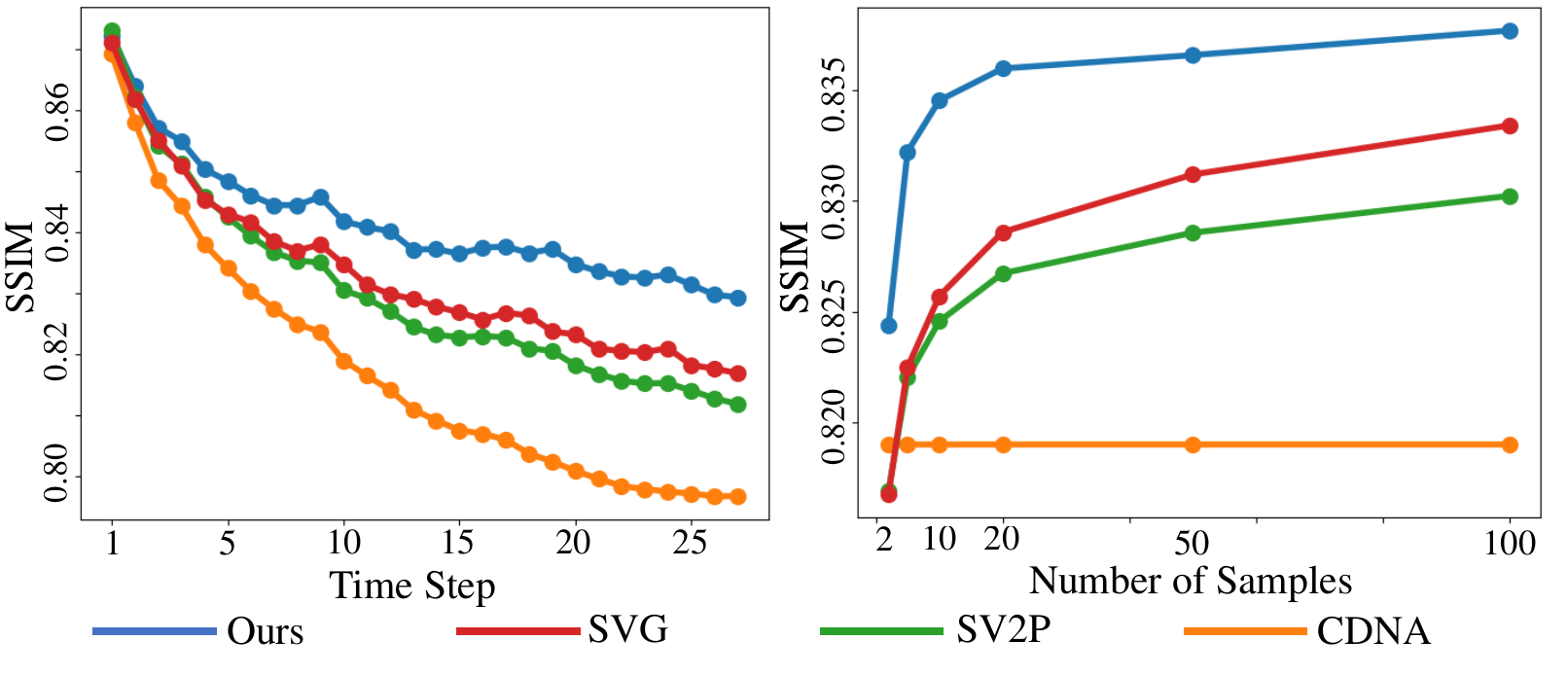}}
\vspace{-0.3cm}
\caption{Evaluation in terms of SSIM on RobotPush~\citep{DBLP:conf/corl/EbertFLL17} dataset. Left figure: X-axis is the time step and Y-axis is SSIM. Right figure: X-axis refers to the number of random samples during evaluation and Y-axis is averaged SSIM over a whole predicted sequence.}
\label{Fig:RobotQuan}
\end{center}
\vspace{-1.0cm}
\end{figure}

\subsection{Robot Arm Motion Prediction}
Experiments on RobotPush~\citep{DBLP:conf/corl/EbertFLL17} dataset take 5 frames as inputs and predict the following 10 frames during training.
As illustrated in Fig.~\ref{Fig:RobotQuan} (first column refers to input), we present quantitative evaluation in terms of SSIM.
For the stochastic method, the best value of 20 random samples is presented.
Fig.~\ref{Fig:RobotQuan} implies that our method outperforms all previous methods by a large margin.
We find CDNA~\citep{DBLP:conf/nips/FinnGL16} (deterministic method) is inferior to stochastic ones. We attribute this to the high uncertainty of robot motion in this dataset. 
Our model, facilitated by example guidance, is capable of capturing the real motion dynamics in a more efficient manner.
\emph{To comprehensively evaluate the distribution modelling ability and sampling efficiency of the proposed method,} we calculate the mean accuracy w.r.t. the number of samples (denoted as $P$):
Fig.~\ref{Fig:RobotQuan} shows the accuracy improvement for all stochastic methods along with the increase of $P$, which tends to be saturated when $P$ is large.
It is worth mentioning that our model still outperforms SVG~\citep{DBLP:conf/icml/DentonF18} by a large margin when $P$ is sufficiently large, e.g., 100. 
This clearly indicates a higher upper bound of accuracy achieved by our model (with guidance of retrieved examples) compared to variational inference based method, i.e., superior capability to capture real-world motion pattern.

\begin{figure*}[t]
\begin{center}
\centerline{\includegraphics[scale=0.34]{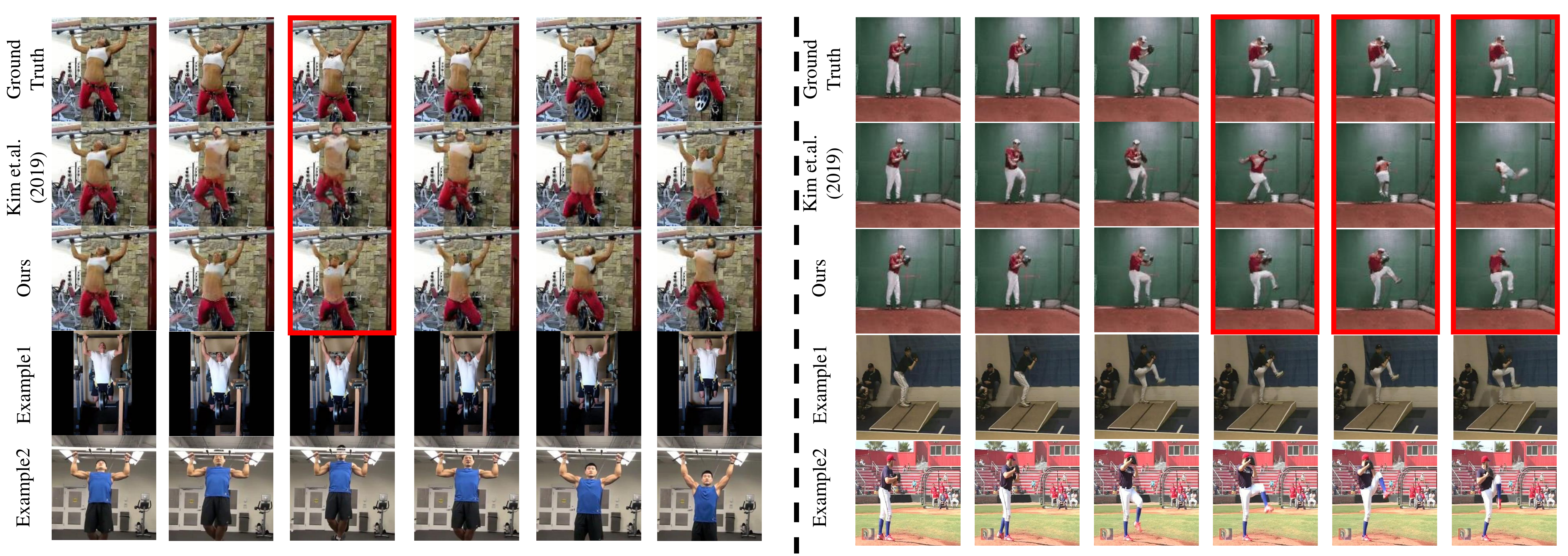}}
\vspace{-0.3cm}
\caption{Qualitative evaluation of human motion prediction on PennAction~\citep{DBLP:conf/iccv/ZhangZD13} dataset. We present ground truth, results of~\citet{DBLP:conf/nips/KimNCK19}, predicted sequence of our model and two searched examples. The left part refers to a pull-up action with multi-modal futures based on the current input and searched examples are capable of matching with possibilities. The right part aims to show that our model is capable of preserving the general structure during prediction. Red-boxes highlight the corresponding evidences on both sides.}
\label{Fig:PennVis}
\vspace{-0.5cm}
\end{center}
\end{figure*}

\begin{table}[b]
  \centering
  \resizebox{\linewidth}{!}{
  \begin{tabular}{c|ccccc}
    \toprule
    \toprule
    Metric & [1] & [2] & [3] & [4] & Ours\\
     \midrule
     Action Acc$\uparrow$ &  15.89 &  40.00 & 47.14 & 68.89 & \textbf{73.23}\\
      FVD$\downarrow$ & 4083.3 & 3324,9 & 2187.5 & 1509.0 & \textbf{1283.5}\\
    \bottomrule
    \bottomrule
  \end{tabular}
  }
  \vspace{-0.3cm}
  \caption{Quantitative evaluation of predicted sequences in terms of Fr\'echet Video Distance (FVD)~\cite{DBLP:journals/corr/abs-1812-01717} (lower is better) and action recognition accuracy (higher is better). Previous works [1]-[4] refer to~\citep{DBLP:conf/eccv/LiFYWLY18,DBLP:conf/icml/WichersVEL18,DBLP:conf/icml/VillegasYZSLL17} respectively. Experiment is conducted on PennAction dataset~\citep{DBLP:conf/iccv/ZhangZD13}.}
  \label{Tab:PennQuan}
  \vspace{-0.7cm}
\end{table}

Predicted sequences are shown in Fig.~\ref{Fig:RobotQual}.
For row arrangement please refer to the caption.
The key region (highlighted with red boxes) of predicted frames is zoomed in for better visualization of details (last column).
Compared to stochastic baselines, our model achieves higher image quality of predicted sequences, i.e., object edges and general structure are better preserved.
Meanwhile, the overall trajectory is more accurately predicted by our model if compared to two stochastic baselines, which is mainly facilitated by the effective guidance of retrieved examples.
For more visual results please refer to the supplementary material.

\begin{table}[t]
  \centering
  \resizebox{\linewidth}{!}{
  \begin{tabular}{c|cccccc}
    \toprule
    \toprule
    Metric & K=2 & K=3 & K=4 & K=5 & K=6 & K=7\\
     \midrule
     PSNR & 17.81 & 18.19 & 18.28 & 18.35 & 18.31 & 18.25\\
     SSIM & 0.78 & 0.82 & 0.83 & 0.84 & 0.84 & 0.83\\
    \bottomrule
    \bottomrule
  \end{tabular}
  }
  \vspace{-0.3cm}
  \caption{Influence of the example number $K$ evaluated in terms of PSNR (first row) and SSIM (second row) on RobotPush~\citep{DBLP:conf/corl/EbertFLL17} dataset. Note that each number reported in this table is averaged over the whole predicted sequence.}
  \label{Tab:N}
  \vspace{-0.5cm}
\end{table}
\subsection{Human Motion Prediction}
We report the experimental results on a human daily activity dataset, i.e., PennAction~\citep{DBLP:conf/iccv/ZhangZD13}.
We follow the setting of~\citet{DBLP:conf/nips/KimNCK19}, which is also a strong baseline for comparison.
More specifically, the class label and first frame are fed as inputs.
Note that under this situation we retrieve the examples according to the first frame in sequences with an identical action label.

\emph{To evaluate the multi-modal distribution modelling capability,} Fig.~\ref{Fig:PennVis}A presents the best prediction sequence of 20 random samples in terms of PSNR and please refer to the caption for row arrangement.
First column refers to input.
The pull-up action generally possesses two motion modalities, i.e., up and down.
We can observe that~\citet{DBLP:conf/nips/KimNCK19} fails to predict corresponding motion precisely even with 20 samples (third time step highlighted with red-boxes).
Our model, guided by similar examples (last two rows in Fig.~\ref{Fig:PennVis}A), is capable of synthesizing the correct motion pattern compared to the groundtruth sequence.
Meanwhile, from Fig.~\ref{Fig:PennVis}B we can notice that~\citet{DBLP:conf/nips/KimNCK19} fails to preserve the general structure during prediction.
The human topology is severely distorted especially at the late stage of prediction (last 3 time steps highlighted with red-boxes).
As comparison the structure of subject is well maintained predicted by our model, which is visually more natural than the results of~\citet{DBLP:conf/nips/KimNCK19}.
This implies reliably capturing the motion distribution facilitates better visual quality of final predicted image sequences.
We present more visual results in supplementary material and please refer to it.

\begin{figure}[t]
\begin{center}
\centerline{\includegraphics[width=\columnwidth]{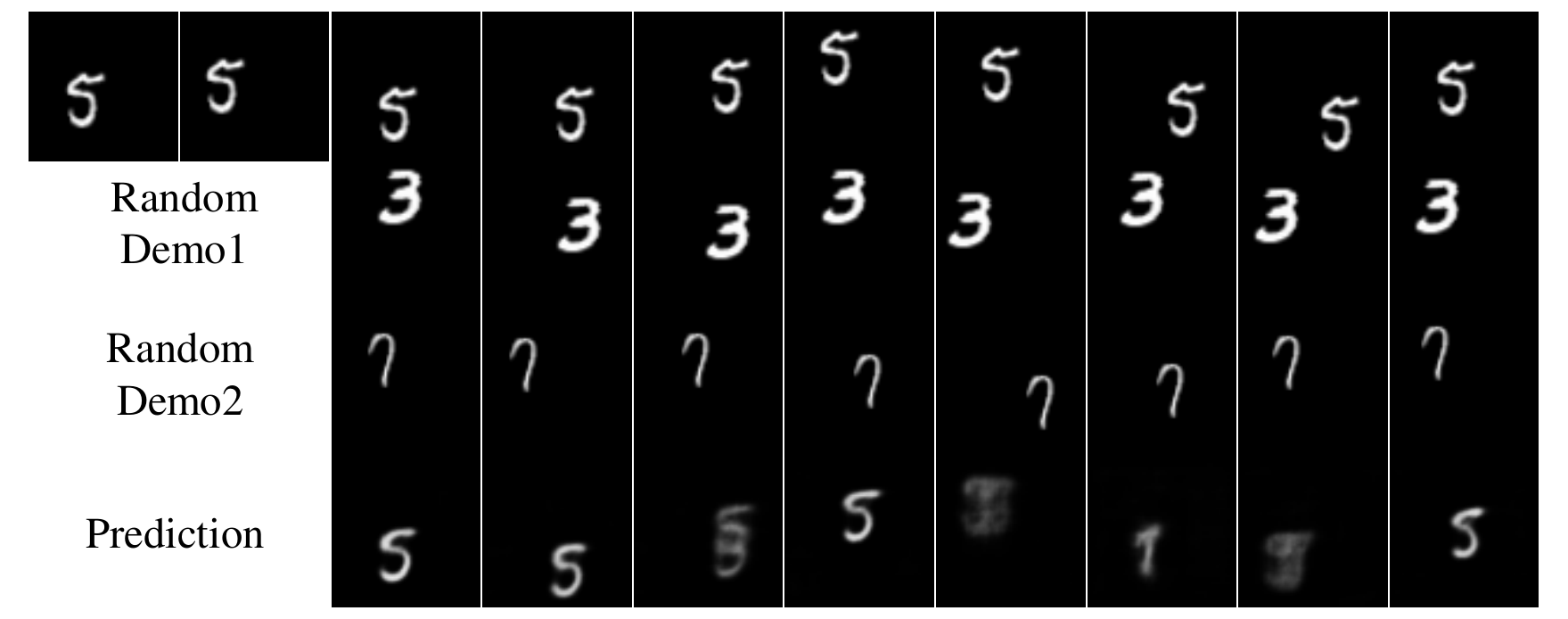}}
\vspace{-0.3cm}
\caption{Prediction results with random example guidance on MovingMnist~\citep{DBLP:conf/icml/SrivastavaMS15} dataset. The top and bottom rows correspond to ground truth and predicted sequence, while the middle two rows are randomly selected examples in this dataset. Unnatural motion is observed during prediction (last row).}
\label{Fig:MnistAbla}
\end{center}
\vspace{-1.0cm}
\end{figure}

For quantitative evaluation, we follow~\citet{DBLP:conf/nips/KimNCK19} to calculate the action recognition accuracy and FVD~\cite{DBLP:journals/corr/abs-1812-01717} score.
As shown in Tab.~\ref{Tab:PennQuan}, our model outperforms all previous methods in terms of both action recognition accuracy and FVD score by a large margin.
This mainly benefits from the retrieved examples, which provides effective guidance for future prediction.

\begin{figure}[t]
\begin{center}
\centerline{\includegraphics[width=\columnwidth]{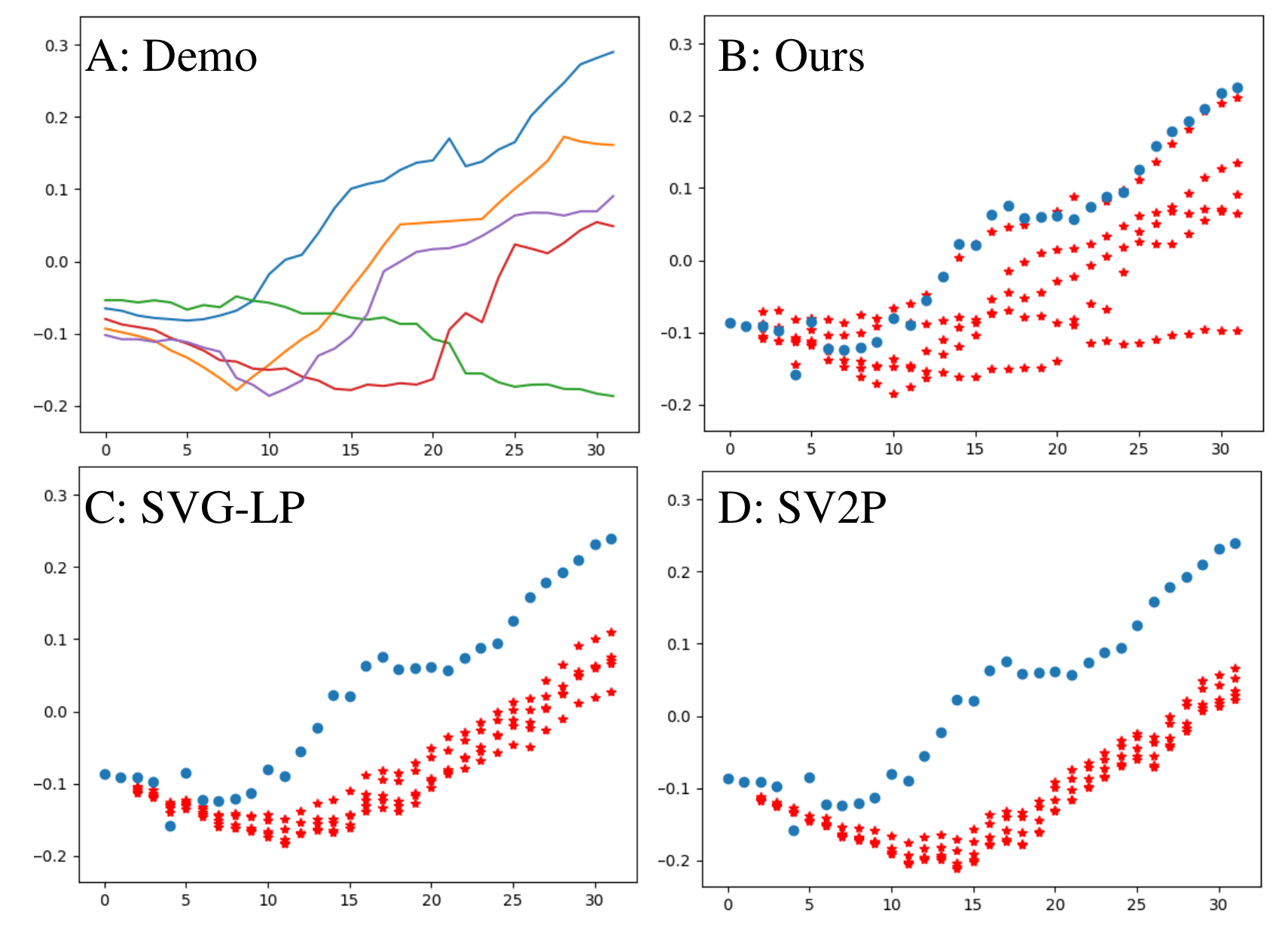}}
\vspace{-0.5cm}
\caption{Visualization of retrieved examples and randomly sampled sequences on RobotPush~\citep{DBLP:conf/corl/EbertFLL17} dataset. Top left refers to searched examples, while the other three figures correspond to sampled sequences by proposed model, SVG~\citep{DBLP:conf/icml/WichersVEL18} and SV2P~\citep{DBLP:conf/iclr/BabaeizadehFECL18} respectively.}
\label{Fig:PennDis}
\end{center}
\vspace{-1.0cm}
\end{figure}

\subsection{Ablation Study}
\textbf{Does example guidance really help?} To evaluate the effectiveness of retrieved examples, we replace the retrieval procedure described in Sec.~\ref{Sec:Demo_Retriv} with the random selection, i.e., the examples have no motion similarity with inputs.
We conduct this experiment on MovingMnist~\citep{DBLP:conf/icml/SrivastavaMS15} dataset.
Results are presented in Fig.~\ref{Fig:MnistAbla} and please refer to caption for detailed row arrangement.
Due to the lack of motion similarity between examples and the input sequence, the predicted sequence demonstrates unnatural motion. The double-image effect of digit 5 (last row in Fig.~\ref{Fig:MnistAbla}), resulting from the misleading information of motion trajectory provided by random examples, implies the critical value of retrieval procedure proposed in Sec.~\ref{Sec:Demo_Retriv}.
In supplementary material, we also present visualization evidence to demonstrate the inferiority of simply combining example guidance and variational inference.
Please refer to it.

\textbf{Does the proposed model really capture multi-modal distribution?}
We present the sampled motion features (Fig.~\ref{Fig:PennDis}) in RobotPush~\citep{DBLP:conf/corl/EbertFLL17} dataset to evaluate the capability of distribution modelling.
For row arrangement please refer to the caption.
For sub-figures from B to D, red-dot lines refer to predicted sequences and blue ones are ground truth.
We can observe that the sampled states of SVG~\citep{DBLP:conf/icml/DentonF18} and SV2P~\citep{DBLP:conf/iclr/BabaeizadehFECL18} are not multi-modal distributed.
Guided by retrieved examples whose multi-modality distribution generally cover the ground truth motion, our model is able to predict the future motion in a more efficient way.
Meanwhile, we present more visualization results to show that predicted sequences are not simply copied from examples and they are highly diverse.

\textbf{Influence of Example Number $K$.}
As illustrated in Tab.~\ref{Tab:N}, we conduct corresponding ablation study about $K$ on RobotPush~\citep{DBLP:conf/corl/EbertFLL17} dataset.
Performance under two metrics, i.e., PSNR and SSIM, is reported.
PSNR and SSIM are averaged over the whole sequence and the best of 20 random sequences is reported.
$K$ ranges from 2 to 7.
We can see that both PSNR and SSIM keep increase when $K$ is no larger than 5 and then decrease.
It indicates that multi-modal examples facilitate better modelling the target distribution, but noise information (or irrelevant motion pattern) might be introduced when $K$ is too large.

\begin{figure}[t]
\begin{center}
\centerline{\includegraphics[width=\columnwidth]{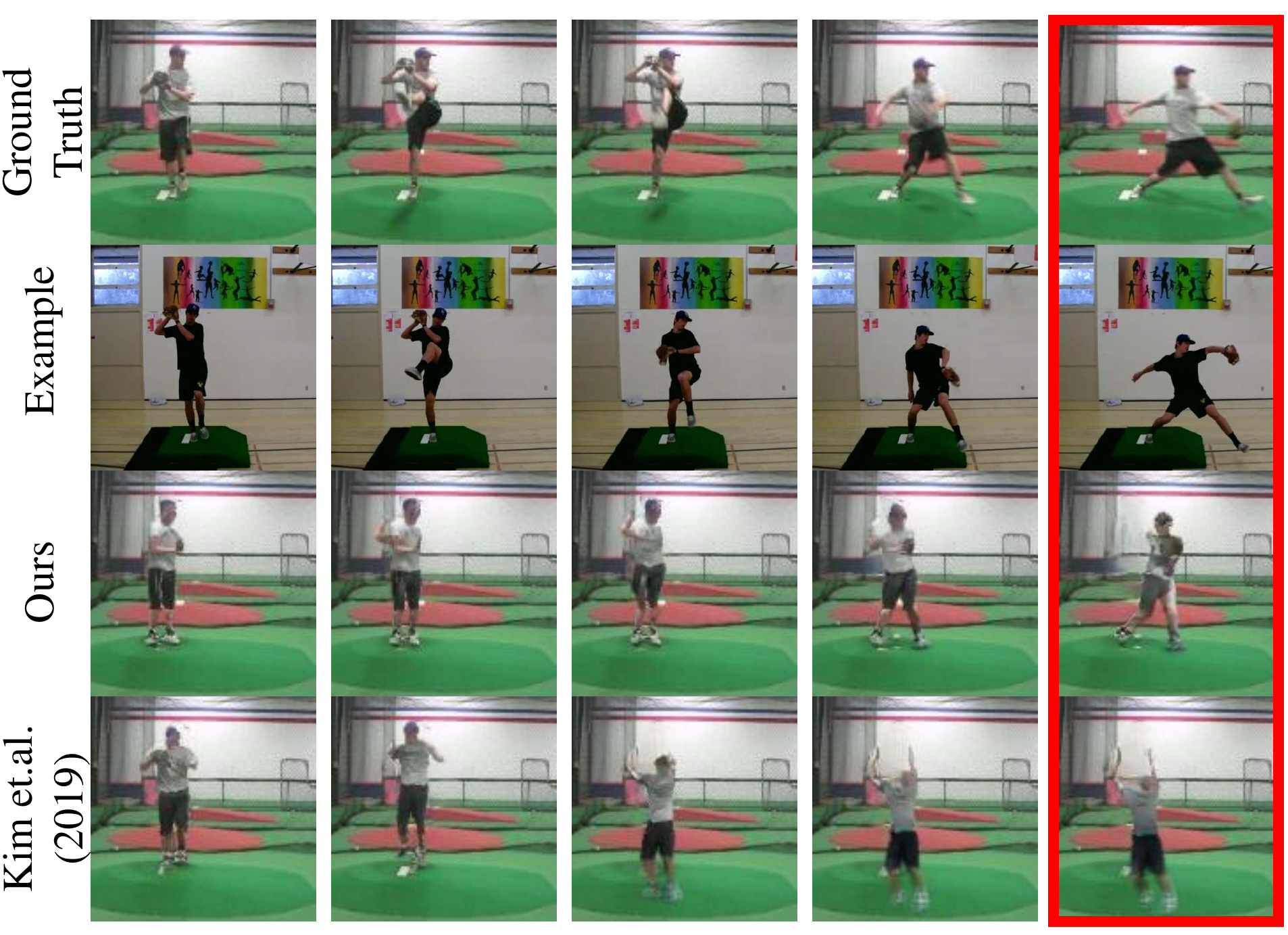}}
\vspace{-0.5cm}
\caption{Predicted results of unseen motion. Contents from top to bottom are ground truth sequence, retrieved example, prediction of our model and the results of~\citep{DBLP:conf/nips/KimNCK19} respectively.}
\label{Fig:Unseen}
\end{center}
\vspace{-1.0cm}
\end{figure}

\subsection{Motion Prediction Beyond Seen Class}
To further evaluate the generalization ability of the proposed model, we are motivated to predict the motion sequence on unseen class.
The majority of video prediction methods are merely able to forecast the motion pattern accessible during training, which are hardly generalizable to novel motion.
We conduct experiments on PennAction~\citep{DBLP:conf/iccv/ZhangZD13} dataset.
We choose three actions, i.e., golf\_swing, pull\_ups and tennis\_serve as known action during training and baseball\_pitch as the unseen motion used during testing.
Our model as well as that of~\citet{DBLP:conf/nips/KimNCK19} is retrained without label class.
During testing, the examples for guidance are retrieved baseball\_pitch sequences.
Fig.~\ref{Fig:Unseen} demonstrates the predicted results.
For row arrangement please refer to the caption.
We can observe that~\citet{DBLP:conf/nips/KimNCK19} fails to give rational prediction regarding the input, where there should be a baseball\_pitch motion but visually resemble tennis\_serve.
Facilitated by the guidance of examples, our model produces a visually natural tennis\_serve sequence, which clearly demonstrates the generalization capability of proposed model.
We argue that the majority of previous works are (implicitly) forced to memorize motion categories in the training set. In contrast to the paradigm, our work is relieved from such burden because the retrieved examples contain the category information in assistance of prediction. We thus focus only on intra-class diversity. 
If given examples with unseen motion categories, our model is still able to give reasonable predictions, thanks to the example guidance.
We present more visual results in supplementary material and please refer to it.

\section{Conclusion}
In this work, we present a simple yet effective framework for multi-modal video prediction, which mainly focuses on the capability of multi-modal distribution modelling. We first retrieve similar examples in the training set and then use these searched sequences to explicitly construct a distribution target. With proposed optimization method based on stochastic process, our model achieves promising performance on both prediction accuracy and visual quality.

\bibliography{example_paper}
\bibliographystyle{icml2020}

\end{document}